\def\year{2019}\relax
\documentclass[letterpaper]{article} 
\usepackage{aaai19}  
\usepackage{times}  
\usepackage{helvet}  
\usepackage{courier}  
\usepackage{url}  
\usepackage{graphicx}  
\usepackage{algorithm}
\usepackage{amssymb}
\usepackage{amsmath}
\usepackage{makecell}
\usepackage{subcaption}
\usepackage{stmaryrd}
\usepackage[noend]{algpseudocode}
\begin{document}
%
\title{Online Risk-Bounded Motion Planning for\\ Autonomous Vehicles in Dynamic Environments}
\author{Xin Huang, Sungkweon Hong, Andreas Hofmann, Brian C. Williams\\
MIT Computer Science and Artificial Intelligence Laboratory \\
32 Vassar St., Cambridge, MA 02139\\
\textit{\{xhuang, sk5050, hofma, williams\}@csail.mit.edu}\\
\thanks{This work was partially supported by the Toyota Research Institute (TRI). However, this article solely reflects the opinions and conclusions of its authors and not TRI or any other Toyota entity.}
}
\maketitle
\begin{abstract}
A crucial challenge to efficient and robust motion planning for autonomous vehicles is understanding the intentions of the surrounding agents. Ignoring the intentions of the other agents in dynamic environments can lead to risky or over-conservative plans. In this work, we model the motion planning problem as a partially observable Markov decision process (POMDP) and propose an online system that combines an intent recognition algorithm and a POMDP solver to generate risk-bounded plans for the ego vehicle navigating with a number of dynamic agent vehicles. The intent recognition algorithm predicts the probabilistic hybrid motion states of each agent vehicle over a finite horizon using Bayesian filtering and a library of pre-learned maneuver motion models. We update the POMDP model with the intent recognition results in real time and solve it using a heuristic search algorithm which produces policies with upper-bound guarantees on the probability of near colliding with other dynamic agents. We demonstrate that our system is able to generate better motion plans in terms of efficiency and safety in a number of challenging environments including unprotected intersection left turns and lane changes as compared to the baseline methods.
\end{abstract}

\section{Introduction}
\label{sec:introduction}

Driving in dynamic scenes, such as intersections and busy streets, is stressful because
drivers need to operate their vehicles safely among other vehicles with uncertain
motions.  A study from \cite{choi2010crash} shows that 40\% of crashes that occurred in the United States in 2008 were intersection-related crashes. Among these intersection crashes, false assumption of other's action and misjudgment of the gap or other's speed are responsible for 8.4\% and 5.5\% of the critical reasons, respectively. These statistics indicate that driving can be safer and less stressful if the intentions of the other vehicles can be recognized accurately. 

Since the first DARPA grand challenge in 2004, there have been many advances in the development of motion planning techniques \cite{paden2016survey} that can be used in intelligent vehicle systems to either assist the driver or take over the control completely. Despite their success, many motion planners work only with static obstacles or assume simple intentions for the dynamic obstacles, such as moving at a constant velocity towards a fixed direction, which can lead to risky or conservative plans. In a recent work \cite{huang2018hybrid}, the authors add uncertainty to the intentions of the moving obstacles such as nearby agent vehicles by assigning a probability distribution over what maneuver each agent vehicle will take, and propose a method that is able to produce risk-bounded motion plans for the ego vehicle that upper bound the probability of near collision with other agent vehicles taking stochastic actions. The work is limited due to its assumption on a fixed distribution over the agent vehicle intentions.

\begin{figure}[t]
\centering
\begin{subfigure}{0.4\textwidth}
\centering
\includegraphics[width=.92\columnwidth]{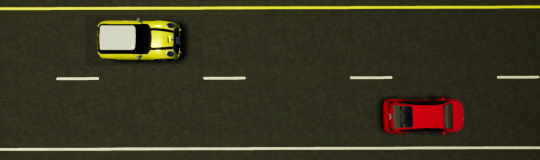}
\end{subfigure}
\begin{subfigure}{0.4\textwidth}
\centering
\includegraphics[width=.92\columnwidth]{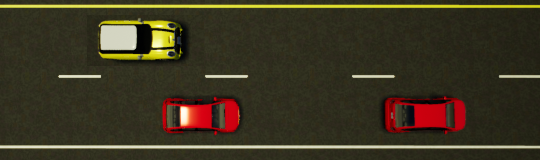}
\end{subfigure}
\caption{Two motivating scenarios, where the ego vehicle in yellow aims to go forward without colliding with the agent vehicle in red. Motion planners that ignore the intentions of the agent vehicles or assume fixed a intention distribution will produce results that are either risky or conservative.}
\label{fig:scenario}
\end{figure}

To motivate this limitation, we show two examples that can be seen in every driving experience in Fig.~\ref{fig:scenario}. In the first case, the autonomous ego vehicle is driving in the left lane behind a slow agent vehicle driving in the right lane. It is a fair assumption that the agent vehicle is more likely going to keep driving in the same lane, so the planner can generate a collision-free plan asking the ego vehicle to go forward and pass the slow vehicle. However, the same assumption fails in the second scenario, where there are two agent vehicles in the right lane, so there is a high likelihood that the following agent vehicle will make a lane change to the left especially when it is faster than the leading vehicle, and the plan of passing it will lead to crash. On the other hand, if we assume that the agent vehicle is always more likely going to make a lane change to the left, the planner will slow down the ego vehicle in the second scenario in order to avoid the crash, but result in the ego vehicle acting too conservatively in the first scenario and failing to make the pass. Therefore, it is critical to dynamically infer the future motion of each surrounding agent vehicle in real-time, which requires a fast intent recognition ability.

We consider the vehicle motion as a sequence of movements depending on the high-level intended maneuver by the driver. Therefore, our proposed intent recognition algorithm is divided into two steps, where we first infer the driver maneuver and then predict a sequence of hybrid vehicle states including discrete maneuver and continuous trajectory over a finite horizon. The discrete maneuver prediction allows the planner to update the probability distribution over the agent vehicle maneuvers, and the continuous trajectory prediction helps to compute the near collision risk and identify risky plans. As studied in \cite{lefevre}, maneuver-based motion prediction approach produces reliable longer-term motion predictions with reasonable computational time as compared to physics-based approaches and interaction-aware approaches.

Due to the uncertainties in the motions of human-driven vehicles, we pre-learn a compact motion representation called Probabilistic Flow Tube (PFT) \cite{dong2012learning} from demonstrating trajectories to capture human-like driver styles and uncertainties for each maneuver. A library of pre-learned PFTs can be used to estimate the current maneuver as well as predict the probabilistic motion of each agent vehicle.

Given the probabilistic prediction results, we can update the belief state dynamically in a POMDP framework and solve it with an existing solver. Compared to many other intention-aware POMDP planners, our method is able to generate risk-bounded plans that provide guarantees on the risk defined as the probability of near collisions with other dynamic agents by augmenting the POMDP model with chance constraints and solving it with a heuristic search algorithm similar to \cite{santana2016rao}. 

Our contributions are three-fold. First, we present a fast and accurate system for recognizing agent vehicle?s intention, in terms of its maneuver and trajectory, using Bayesian filtering and a set of pre-learned maneuver motion models that consider maneuver uncertainties and encode different driver styles. Second, we present an intention-aware conditional motion planner that produces risk-bounded plans in dynamic driving scenes by considering the stochastic motions of other dynamic agents and the upper bound on the risk constraints. Lastly, we show that our planner can produce policies that echo different driver styles in an uncertain environment by adjusting the upper bound probability for the risk constraints in the POMDP model.

\section{Related Work}
\label{sec:related_work}
In this section, we discuss state-of-the-art works relevant to our work in three areas: motion planning, motion prediction, and motion representation.

\subsection{Motion Planning with Moving Obstacles}

Vehicle motion planning problems with moving obstacles have been studied for decades. Many approaches use search-based methods to find a feasible path to the goal location by considering future motions of moving obstacles \cite{rudenko2017predictive,bansal2018collaborative,ajanovic2018search}. However, these methods rely on simple prediction models, such as vehicle driving at current velocity and staying in the same lane \cite{ajanovic2018search}, and goal states of the obstacles are known a priori \cite{rudenko2017predictive,bansal2018collaborative}. Such models are not realistic in planning for long-horizon where the obstacles can change both high-level goals and low-level paths over time.

A widely used framework for motion planning with dynamic obstacles is called Partially Observable Markov Decision Process (POMDP), which provides a systematic model that incorporates the uncertainty in the environment including the sensor noise and stochastic future motions of surrounding obstacles. Approaches in \cite{bandyopadhyay2013intention,ulbrich2013probabilistic,brechtel2014probabilistic,bai2015intention,luo2018porca} model the intention-aware motion planning problem with POMDP where the intentions of the dynamic obstacles, such as vehicles and pedestrians, are represented by a belief state, and solve for the POMDP instance using approximation methods due to the complexity of the search space.  Instead of solving the complete POMDP instance directly, \cite{chen2016pomdp} decouples the problem into multiple MDP instances by assuming fixed or deterministic intention. Additionally, Bayesian reinforcement learning \cite{wang2012monte,hoang2013general} and deep reinforcement learning \cite{everett2018motion} are proposed to enable safe and efficient motion planning in dynamic environments. These methods are limited in either working in discrete space or requiring a large amount of training data and computational resources.

Despite the success of POMDP-based approaches, they fail to provide a plan with a guarantee over the probability of success, such as avoiding near collision with moving agents in a dynamic environment. An extension called constrained POMDP is introduced to model risk explicitly and generate plans with bounded-risks as the world changes \cite{undurti2010online,poupart2015approximate}. Using a similar idea, a chance-constrained POMDP (CC-POMDP) is proposed to include a more flexible definition of risk allocation and studied in different problem domains \cite{santana2016rao}. Unfortunately, the model assumes fixed or deterministic intentions of the dynamic agents, which can fail in many realistic driving scenarios. Therefore, we decide to improve the chance-constrained model by updating the belief state dynamically with a real-time intent recognition system, and solve it by a heuristic search algorithm named Risk-Bounded AO* (RAO*) \cite{santana2016rao}.

\subsection{Maneuver-based Motion Prediction}

Maneuver-based prediction algorithms predict the future trajectory of a target vehicle by estimating and predicting a sequence of high-level maneuver actions. 

Vehicle maneuvers can be estimated and predicted using many approaches including threshold-based classifiers \cite{houenou}, Bayesian Networks \cite{schreier2014bayesian}, Hidden Markov Model \cite{li}, logistic regression \cite{klingelschmitt2014combining}, Gaussian process regression \cite{tran2013modelling}, and deep neural networks \cite{deo2018multi}. In this work, we focus on methods that output probability distributions over both discrete maneuvers and continuous trajectories.

The maneuver estimation results can help predict continuous trajectories over longer horizons. For example, \cite{houenou} generates a set of long-term trajectories based on the detected maneuver and selects the one with the minimum cost. In \cite{deo2018multi}, the authors train a deep neural network to generate a probability distribution over future locations conditioned on the states of the vehicle and the estimated maneuver. In \cite{schreier2014bayesian}, a hand-crafted motion model is used to predict future trajectories given the estimated maneuver. Similarly, our method uses a pre-learned motion model to generate probabilistic predictions on the target vehicle's future positions.

\subsection{Vehicle Motion Representation}
Different motion representations have been used in the vehicle motion prediction domain. Examples include poly-lines \cite{houenou}, maneuver-specific models \cite{schreier2014bayesian}, Gaussian mixture models \cite{deo2018multi}, and funnels \cite{tobenkin2011invariant}, etc. Since some trajectories are more likely to occur than others when a human driver executes a particular maneuver, we favor probabilistic motion representations, which are useful to incorporate the uncertainties from human driver behavior and to compute the near collision probability in a chance-constrained framework. 

The work in \cite{dong2012learning} proposes a representation called Probabilistic Flow Tubes that models human-like actions in humanoid robotics tasks by computing common characteristics defined over a time interval from a set of demonstrating examples. 

We decide to use PFT to model probabilistic vehicle motions because of a few reasons. First, it is a compact representation to capture both temporal and spatial aspects of the maneuver motions, which is parameterized by a sequence of Gaussian distributions defined by means and covariances over different time steps. Second, it learns human behavior through a number of demonstrating examples. In the vehicle domain, the PFT can capture different driving styles efficiently given enough demonstrating trips. Although many approaches estimate the driving styles more explicitly from onboard vehicle sensors \cite{martinez2018driving}, PFT is able to incorporate driver styles inherently in its representation.

\section{Problem Formulation}
\label{sec:formulation}
\begin{figure}[t!]
\centering
\includegraphics[width=0.96\columnwidth]{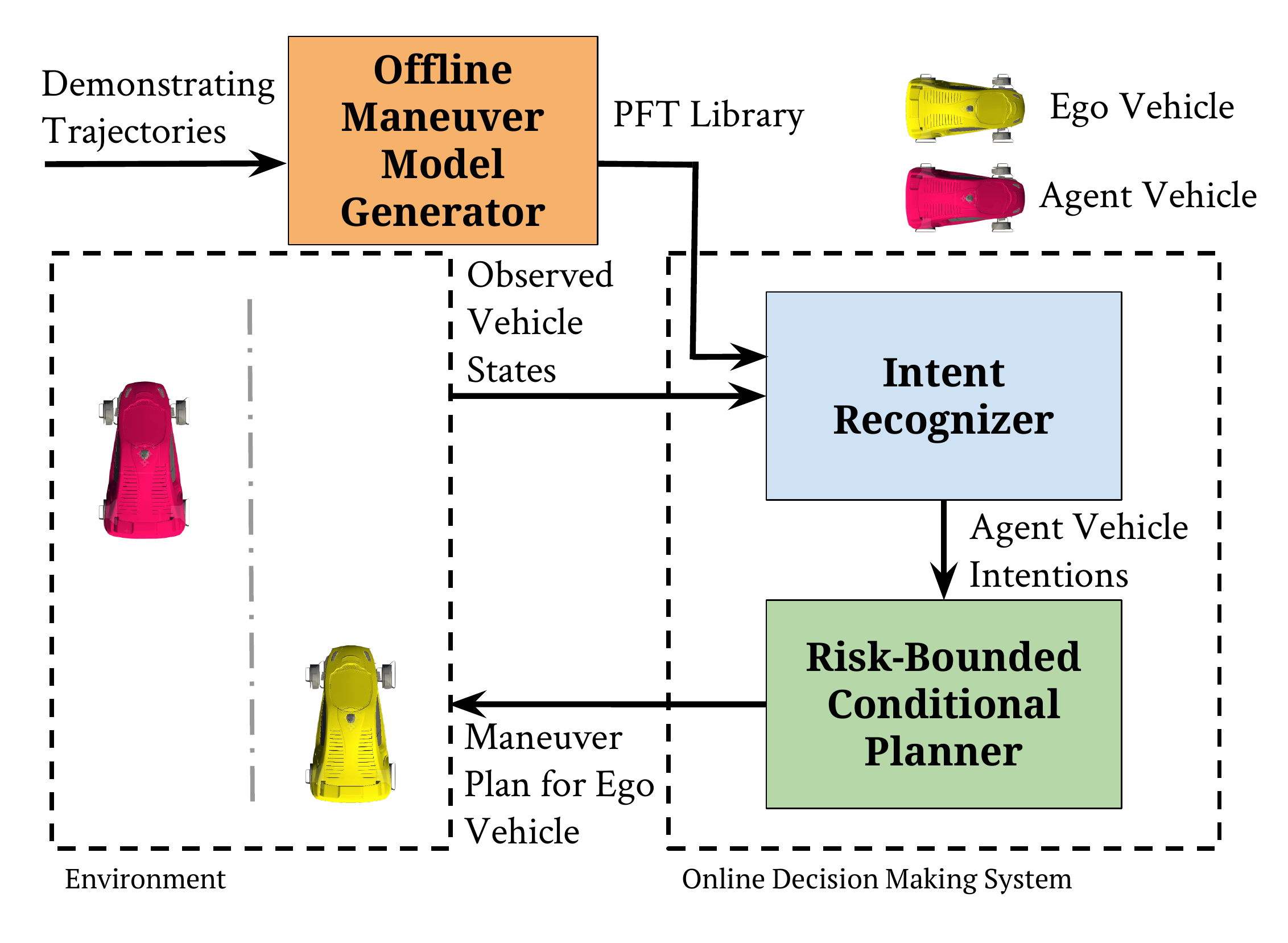}
\caption{Architecture diagram for our risk-bounded intention-aware motion planning system.}
\label{fig:architecture}
\end{figure}

Our goal is to generate a motion plan for a controllable ego vehicle that navigates in a dynamic environment with other uncontrollable agent vehicles with stochastic motions. The input includes a library of pre-learned maneuver motion models and a sequence of observed vehicle states. The output is a policy that achieves the minimum cost while maintaining an upper bound on risk defined as the probability of near colliding with other agent vehicles. The overall architecture diagram of our proposed planning system is depicted in Fig.~\ref{fig:architecture}.

We start by modeling the vehicle motion planning problem with a \textit{CC-POMDP} instance  \cite{santana2016rao} that defines the problem of planning with multiple agents including a controllable agent $\mathcal{R}_0$ and a number of $n$ uncontrollable agents $\mathcal{R}_1, \ldots, \mathcal{R}_n$ with no communication in between. The vehicle-specific instance is a tuple $\mathcal{P} = \langle \mathcal{S}, \mathcal{A}, \mathcal{O}, T, O, C, \mathcal{B}_0, H, \Delta \rangle$, where
\begin{itemize}
\item $\mathcal{S}$ is a set of hybrid states for each vehicle including the positions and intended maneuvers, where the ego vehicle has no intended maneuvers.
\item $\mathcal{A}$ is the set of maneuvers that can be taken by the ego vehicle.
\item $\mathcal{O}$ is a set of observations on the vehicle positions.
\item $T: \mathcal{S} \times \mathcal{A} \times \mathcal{S} \to \mathbb{R}$ is a state transition function between hybrid states. The transition function depends not only on the action chosen by $\mathcal{R}_0$, but also on the intention of each uncontrollable vehicle.
\item $O: \mathcal{S} \times \mathcal{O} \to \mathbb{R}$ is an observation function that models the probability of observing a sequence of agent vehicle positions given the state information such as intended maneuvers.
\item $C: \mathcal{S} \times \mathcal{A} \to \mathbb{R}$ is a cost function associated with each action in $\mathcal{A}$.
\item $\mathcal{B}_0$ is the initial belief state over hybrid states.
\item $H$ is the finite planning horizon.
\item $\Delta$ is an upper bound probability for the risk constraint, where the risk is defined as the probability of $\mathcal{R}_0$ near colliding with any of the agent vehicles up to the planning horizon. 
\end{itemize}

Given the instance, the objective is to find an optimal, deterministic, and chance-constrained policy $\pi^*$ at time step $k$, such that
\begin{equation}
\pi^* = \arg \min_{\pi} \mathbb{E} \Bigg[ \sum_{t=k}^{k+H} C(s_t, a_t) \Big| \pi \Bigg],
\end{equation}
subject to
\begin{equation}
\label{eq:er}
er(\mathcal{B}_k | \pi^*) = Pr(\bigvee_{i=k}^{k+H} \text{Collision}_i|\mathcal{B}_k,\pi^*) \leq \Delta,
\end{equation}
where $er$ is the execution risk of a policy $\pi$ measured from the current belief state $\mathcal{B}_k$ over planning horizon $H$.

We assume the intended maneuvers of the agent vehicles to be partially observable, and introduce a Probabilistic Hybrid Automaton (PHA) including as a 6-tuple $\mathcal{H}= <\mathbf{x}, \mathbf{w}, \mathit{F}, \mathit{T_d}, \mathcal{X}_{d}, T_{s}>$ \cite{hofbaur2002mode} to estimate the hybrid states for each uncontrollable agent vehicle:

\begin{itemize}
    \item $\mathbf{x} = \mathbf{x}_{d} \cup \mathbf{x}_{c}$ are the state variables, where $\mathbf{x}_{d} \in \mathcal{X}_{d}$ includes an intended maneuver type and a maneuver clock, and $\mathbf{x}_{c}$ denotes continuous state variables as global positions $(x, y)$ of the target vehicle.
    \item $\mathbf{w} = \mathbf{u}_{d} \cup \mathbf{o} \cup \mathbf{v}_{c}$ are the interface variables, where $\mathbf{u}_{d} = \emptyset$ denotes the set of discrete input variables because we assume no access to the control over the uncontrollable agents,  $\mathbf{o} = \{z_x, z_y\}$ includes the observed vehicle locations, and $\mathbf{v}_{c} = \emptyset$ assumes the vehicle positions are fully observable.
    \item $\mathit{F} = \mathcal{F}_{c} \cup \mathcal{G}_{c}$ specifies the continuous evolution of the automaton. For each discrete state $x_{d} \in \mathcal{X}_{d}$, $F_{c}$ specifies the discrete-time dynamic evolution of the vehicle, and is represented by a Probabilistic Flow Tube (PFT) with sampling period $T_{s}$. More details of PFT will be explained in Section \textsc{Approach}. In addition, $\mathcal{G}_{c}(a) = a$ is a perfect observation function for any continuous variable $a$.  
    \item $\mathit{T_d}$ specifies a set of transition functions for each discrete state. Each transition function has an associated guard condition determined by the continuous state variables. 
\end{itemize} 

After defining a \textit{PHA} specifically for each uncontrollable vehicle, we introduce a sub-problem of intent recognition as generating the probability distribution over the hybrid state over a finite predicting horizon $h$ that consists of continuous trajectory $\mathbf{x}_c$ together with the discrete state $\mathbf{x}_d$ given a sequence of observed history vehicle locations $\mathbf{o}_{k-w+1:k}$ with length $w$: 
\begin{equation}
P(\mathbf{x}_{c,k:k+h},\mathbf{x}_{d,k:k+h}|\mathbf{o}_{k-w+1:k}).
\end{equation}

Note that the predicting horizon $h$ is different from the planning horizon $H$ defined in the POMDP model. The predicting horizon refers to the number of time steps ahead of the current time step for intent recognition, and the planning horizon refers to the number of actions the planner needs to look ahead. Usually, the predicting horizon is the product of the planning horizon and the number of clock cycles contained in the maneuver action. 

Finally, the intent recognition results can be used to update the belief state and the transition function $T$ in the \textit{CC-POMDP} instance, as explained in the next section.

\section{Approach}
\label{sec:approach}
This section introduces three major modules in our intention-aware motion planning system as illustrated in Fig.~\ref{fig:architecture}: a \textit{Motion Model Generator} generating a library of Probabilistic Flow Tubes (PFTs) that model the continuous evolution of high-level vehicle maneuvers, an \textit{Intent Recognizer} that predicts probabilistic maneuver states and continuous vehicle positions over a finite horizon for each uncontrollable agent vehicle, and a \textit{Risk-bounded POMDP Planner} that considers the agent car intentions and produces the desired maneuver plan for the ego car.

\subsection{Probabilistic Maneuver Motion Modeling}
\textit{Probabilistic Flow Tubes} (PFTs) are used to model the probabilistic continuous motion for autonomous systems. As described in \cite{dong2012learning}, a PFT is a compact representation for continuous trajectories with common characteristics. It is parameterized by a set of means and covariances that correspond to each time step of the common trajectories. 

Since the trajectories are generated from human demonstrations, PFTs are useful to model realistic human behaviors and incorporate different driver styles. The algorithm to generate a PFT for each discrete maneuver $M \in \mathcal{M}$ given a set of demonstrating trajectories $\mathcal{D}_M$ is shown in Alg. \ref{alg:pft}. Since the trajectories associated with the same maneuver may have different lengths, we first perform dynamic time warping (DTW) on the input trajectories to temporally align them \cite{myers1980performance}. Assuming all aligned trajectories have a length of $L$, we compute the means and covariances among all aligned trajectories at time steps from $1$ to $L$.

\begin{algorithm}[H]
    \caption{Generate a Probabilistic Flow Tube for $M$}\label{euclid}
     \label{alg:pft}
   \hspace*{\algorithmicindent} \textbf{Input} $M, \mathcal{D}_M = \{D_M^{(n)}\}_{n=1\ldots N}$,  a set of $N$ trajectories\\
    \hspace*{\algorithmicindent} \textbf{Output} $L_M$
    \begin{algorithmic}[1]
    \Procedure{GeneratePFT}{$\mathcal{D}_M$}
    \State $\mathcal{D}'_M$ = DTW($\mathcal{D}_M$)
    \For{$i=1$ to $L$}
    	\State $\mathbf{x}_i \gets (D^{'(1)}_{M}[i], \ldots, D^{'(N)}_{M}[i])$
	\State $\mu[i] = \textrm{mean}(\mathbf{x}_i)$
	\State $\Sigma[i] = \textrm{covariance}(\mathbf{x}_i)$
    \EndFor
    \State $L_M = (\mu, \Sigma)$
    \State \textbf{return} $L_M$
    \EndProcedure
    \end{algorithmic}
\end{algorithm}

Therefore, given a discrete state including a maneuver label $M$ and a maneuver clock $i$, the probability of the continuous state is given as a Gaussian distribution:
\begin{equation}
    p(\mathbf{x}_c | \mathbf{x}_d = (M, i)) = \mathcal{N}(\mu_M[i], \Sigma_M[i]).
\label{eq:pft_stats}
\end{equation}

In addition, similar to \cite{dong2012learning}, the likelihood of observing a continuous past trajectory $\mathbf{o}_{k-w+1:k}$ given a maneuver label $M$ and a maneuver clock $i$ is the product of probability densities of each Gaussian distribution in the flow tube evaluated at the aligned points in the observed vehicle trajectory. Since the trajectory is not necessarily aligned with the flow tube, we shift the end of the trajectory so that it is aligned with $\mu_M[i]$ and name the shifted trajectory $o'_{k-w+1:k}$. Therefore, the likelihood is computed as:
\begin{multline}
p(\mathbf{o}_{k-w+1:k} | \mathbf{x}_d = (M, i)) \\
= \prod_{j=k-w+1}^k \mathcal{N}(\mathbf{o}'_{j} | \mu_M[i-k+j], \Sigma_M[i-k+j]).
\label{eq:pft_pdf}
\end{multline}

\subsection{Probabilistic Hybrid State Prediction}
We divide the probabilistic hybrid predictor into two steps. In the first step, we estimate the discrete maneuver state at the current time step $k$ using Bayesian filtering and a library of pre-learned PFTs. Next, given the discrete estimation result, we compute the probability distribution over the future hybrid states using a recursive approach.

\subsubsection{Discrete State Estimation}
Estimating the hybrid state in a \textit{PHA} is considered as a \textit{hybrid mode estimation problem} \cite{hofbaur2002mode}. Since the continuous vehicle locations are assumed to be fully observable, our goal is simplified to generate the probability distribution over the discrete maneuver state $\mathbf{x}_d$ at the current time $k$ given a \textit{PHA} and an observation sequence $\mathbf{o}_{k-w+1:k}$ (simplified as $\mathbf{o}$) with length $w$.

The probability distribution can be computed by multiplying the observation probability using Equation \ref{eq:pft_pdf} and the prior over the discrete states, where the prior is initialized to be a uniform distribution:
\begin{equation}
p(\mathbf{x}_{d,k} | \mathbf{o}) \propto p(\mathbf{o} | \mathbf{x}_{d,k}) p(\mathbf{x}_{d,k}).
\label{eq:filter}
\end{equation}

\subsubsection{Hybrid State Prediction}
Given the estimation on the discrete state at current step $k$, we wish to predict the future hybrid vehicle states over a predicting horizon of $h$. To start, we compute the probability distribution over the next hybrid step at $k+1$:
\begin{equation}
p(\mathbf{x}_{d,k+1}, \mathbf{x}_{c,k+1} | \mathbf{o}) = p(\mathbf{x}_{d,k+1} |\mathbf{o}) p(\mathbf{x}_{c,k+1} | \mathbf{x}_{d,k+1}, \mathbf{o}) 
\label{eq:pred}
\end{equation}

The first term $p(\mathbf{x}_{d,k+1} |\mathbf{o})$ in Equation \ref{eq:pred} can be computed by Equation~\ref{eq:pred1}, where $p(\mathbf{x}_{d,k+1} |\mathbf{x}_{d,k}, \mathbf{o})$ can be computed using the discrete transition function $T_d$ and its associated guard condition determined by observations $\mathbf{o}$, as specified in the \textit{PHA} model, and $p(\mathbf{x}_{d,k} |\mathbf{o})$ is given by Equation \ref{eq:filter}:
\begin{equation}
p(\mathbf{x}_{d,k+1} |\mathbf{o}) = \sum_{\mathbf{x}_{d,k}} p(\mathbf{x}_{d,k+1} |\mathbf{x}_{d,k}, \mathbf{o}) p(\mathbf{x}_{d,k} |\mathbf{o}). 
\label{eq:pred1}
\end{equation}

The second term $p(\mathbf{x}_{c,k+1} | \mathbf{x}_{d,k+1}, \mathbf{o})$ in Equation \ref{eq:pred} represents the probability distributions of vehicle locations given the predicted discrete maneuver state, which can be computed by looking up the PFT associated with the discrete maneuver state using Equation \ref{eq:pft_stats}.

Applying Equations \ref{eq:pred} and \ref{eq:pred1} recursively, we can estimate the  future maneuver sequences as well as continuous states over a predicting horizon of $h$ for the target vehicle.

The size of possible values for $\mathbf{x}_{d,k:k+h}$ increases exponentially with $h$, which causes the POMDP solver to search over a very large search space. To reduce the size of possible discrete states, we set the probability of very low likely events (e.g., with probability smaller than a pre-defined threshold $\epsilon$) at each predicting step to be 0. 

\subsection{Intention-Aware Risk-Bounded Motion Planning}
In this section, we present an online intention-aware planning algorithm by combining a heuristic forward search method with intent recognition results, as outlined in Algorithm \ref{alg:intentionrao}.

\begin{algorithm}[H]
    \caption{Online Intention-Aware Motion Planning}
    \label{alg:intentionrao}
    \textbf{Input} CC-POMDP $\mathcal{P}$, \textit{PHA} $\mathcal{H}$, library of PFTs $\mathcal{L}$. \\
    \textbf{Output} None.
    \begin{algorithmic}[1]
    \State $\mathcal{P}_0 \gets \mathcal{P}$
    \State $\mathcal{B}_0 \gets$ GetInitialBeliefState$(\mathcal{P}_0)$
    \State $k \gets 1$
    \While {True}
    	\State $\mathbf{o}_k \gets$ ObserveCurrentVehicleLocations()
    	\State $\mathbf{o}_{k-w+1:k} \gets$ UpdateObservation($\mathbf{o}_k, \mathbf{o}_{k-w:k-1}$)
    	\State $\mathbf{p}(\mathbf{x}_{d,k}) \gets$ DiscreteEstimation($\mathbf{o}_{k-w+1:k}, \mathcal{H}, \mathcal{L}, \mathcal{B}_{k-1}$)
        \State $\mathcal{B}_{k} \gets$ UpdateBeliefState($\mathbf{o}_k, \mathbf{p}(\mathbf{x}_{d,k})$)
        \If {AtGoal($\mathcal{B}_{k})$}
            \State Output success.
            \State \textbf{break}
        \EndIf
        \State $\mathbf{p}(\mathbf{x}_{d\shortrightarrow}, \mathbf{x}_{c\shortrightarrow}) \gets$ Predict($\mathbf{o}_{k-w+1:k}, \mathcal{H}, \mathcal{L}, \mathbf{p}(\mathbf{x}_{d,k})$)
        \State $T_k \gets$ UpdateTransition($\mathbf{p}(\mathbf{x}_{d\shortrightarrow}, \mathbf{x}_{c\shortrightarrow})$)
        \State $\mathcal{P}_k \gets$ UpdateModel($\mathcal{P}_{k-1}, \mathcal{B}_k, T_k$)
        \State $\pi_k \gets $ RAO\text{*}($\mathcal{P}_k$)
        \State Execute(GetAction($\pi_k$, $\mathcal{B}_{k}$))
        \State $k \gets k+1$
    \EndWhile
    \end{algorithmic}
    \label{alg:planner}
\end{algorithm}

We first initialize the necessary variables given the inputs to the problem (Line 1-3). At each execution step $k$, we begin by updating the observation sequence by collecting the latest vehicle positions through perception sensors (Line 5-6). We assume that we always have enough observations from the past. Since the major uncertainty in this work comes from the intentions of agent vehicles, we assume that the perception sensors are perfect and thus the vehicle states are fully observable. 

Next, we estimate the possible maneuvers for each agent vehicle by taking the prior from the current belief state and computing the posterior using Equation \ref{eq:filter} (Line 7). The estimation results and the most recent observed vehicle positions are used to update the current belief state (Line 8). If the belief state satisfies the goal condition, such as the ego vehicle arriving at a predefined location or region, we output success and end the planning process. Otherwise, we need to continue to predict the future hybrid states $\mathbf{p}(\mathbf{x}_{d, k:k+h}, \mathbf{x}_{c, k:k+h})$ (abbreviated as $\mathbf{p}(\mathbf{x}_{d\shortrightarrow}, \mathbf{x}_{c\shortrightarrow})$) for all agent vehicles (Line 12). The predictions are incorporated into the transition function, which keeps information on how the agent vehicles will move in the future (Line 13-14).

Once the POMDP is updated with the latest intent recognition results, our system can utilize any off-the-shelf POMDP solver to find a feasible policy. Because of the bounding risk requirement in our problem, we use a method called risk-bounded AO* (RAO*) \cite{santana2016rao,huang2018hybrid} that finds the solution through heuristic forward search for CC-POMDP (Line 15), where the value heuristic is defined as the Euclidean distance to the goal location. To search for a risk-bounded policy, RAO* uses a second risk heuristic, based on execution risk, to prune the search space by removing the over-risky branches that violate the risk constraints. The execution risk, as defined in Equation~\ref{eq:er}, is computed as the probability of near collisions between our ego vehicle and agent vehicles, based on the intent recognition results. 

In order to find solutions in real-time, progress has been made to speed up RAO* including adding preconditions for each action to further prune the search space and considering only the interaction with nearby vehicles to reduce the branching factor. Another possible way to improve computational time is to use an iterative RAO* algorithm that reuses the search graph during replanning \cite{huang2018hybrid}. Unfortunately, the changing intentions of the agent cars restrict the amount of search graph that can be reused, which means in many cases we need to build a new graph from scratch. Therefore, for the sake of simplicity, RAO* is used and we show in the next section that it is sufficient to solve the vehicle motion planning problem with short horizons.

Given the output policy from the solver, our system can quickly look up the optimal action to execute (Line 16). The system would continue to iterate through the same steps in a loop until the goal condition is satisfied.

In general, the online intention-aware planning ability allows our system to make adjustments quickly and always have a risk-bounded plan when the surrounding agent vehicles change their intentions. This is a major advantage of our system as compared to other works using offline policies or assuming a fixed distribution over the vehicle intentions.

\section{Experiments}
In order to evaluate our system, we create an intersection unprotected left turn scenario and a dynamic lane change scenario in a state-of-the-art urban driving simulator called CARLA \cite{Dosovitskiy17}. The simulator is able to produce a realistic driving experience in a small town and allows the users to create multiple vehicles based on their needs. Each vehicle can be controlled individually through steering wheel angle and throttle inputs collected from either a user input device such as a keyboard or a programmed software that generates the control given the environment information. In each scenario, we aim to test the different aspects of our system, as detailed in the rest of this section.

\subsection{Intersection Left Turn} 
According to \cite{choi2010crash}, intersections are responsible for 40\% of crashes occurred in the United States. Among all intersection crashes, unprotected left turns are one of the most common and dangerous ones. Therefore, we wish to verify our intention-aware planning system in a simplified scenario, where the ego car needs to make a left turn at an unprotected intersection when an agent car is approaching from the opposite direction. We aim to show that our system is able to handle this dynamic environment and outperform other baseline methods. A follow-up experiment demonstrates that our system is able to simulate different driving styles by tuning the upper bound on the risk-constraint.

\subsubsection{Scenario Setup}
\begin{figure}[t!]
    \centering
    \begin{subfigure}[b]{0.24\textwidth}
        \centering
        \includegraphics[height=1.4in]{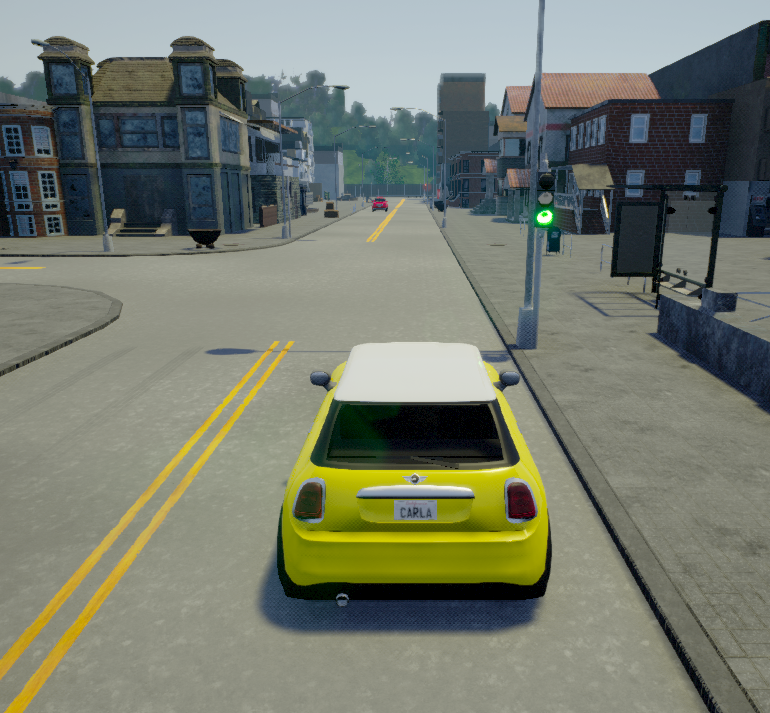}
        \caption{View of the ego car.}
        \label{fig:ego_view}
    \end{subfigure}%
    \begin{subfigure}[b]{0.24\textwidth}
        \centering
        \includegraphics[height=1.4in]{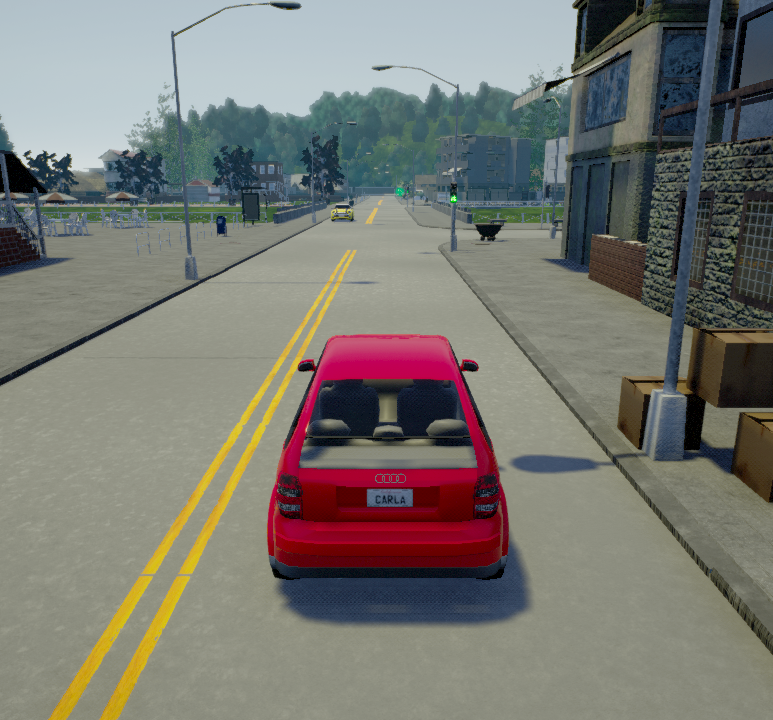}
        \caption{View of the agent car.}
        \label{fig:agent_view}
    \end{subfigure}
    \caption{Intersection left turn scenario setup.}
    \label{fig:left_turn_setup}
\end{figure}

As shown in Fig.~\ref{fig:left_turn_setup}, we start the test by placing the yellow ego vehicle in front of an intersection, and the red agent vehicle on the opposite side. The ego car has a goal to turn left. At each step, it can choose to stop and yield to the agent car or perform a left turn. In order to encourage turning sooner than later, we add a higher cost to the stop action. On the other hand, the agent car chooses randomly from two maneuvers including going forward or slowing down to yield to the ego car. We pre-learn a library of PFTs representing the uncertainties of these two maneuvers by asking several human drivers to teleop the agent car for 100 times in the simulator using a Logitech steering wheel and pedal device. Furthermore, to maintain a reasonable computational time, we choose to use 0.0001 as the value of $\epsilon$ that filters out approximately 90\% of possible discrete states. In addition, as described in the definition of \textit{PHA}, we assume that the ego vehicle has access to the exact locations of all agent vehicles.

\subsubsection{Results}
We start by investigating the performance of our intent-recognition algorithm by separately measuring the accuracy of maneuver estimation and trajectory prediction. To measure the maneuver estimation accuracy, we compute the percentage of correctly identified maneuvers among the testing set. Since the intent recognition algorithm outputs a probability distribution over possible maneuvers, we choose the most likely maneuver and compare it with the ground truth maneuver. To measure the motion prediction performance, we sample from the predicted trajectory distribution and compute the average end displacement error over 4.8 seconds among all testing samples. The predicting horizon is chosen as the time taken by the ego car to make a left turn. Among all the testing examples, our intent estimation has an average accuracy of 96.19\%, and the average end displacement error is 2.02 meters. We observe that most maneuver misclassfications are due to some forward testing trajectories that contain a sequence of consecutive positions connected with small gaps. Such sequence confuses the maneuver estimator to produce a distribution that assigns more probability to the slow down maneuver than to the forward maneuver, as well as causes a large trajectory prediction error.

Next, we want to check the performance of our proposed intention-aware motion planning approach. Since this work focuses on extending a risk-bounded motion planner with intention recognition capabilities, we compare our approach with three baseline planners that have simple prediction models. The first baseline planner assumes that the agent car will choose each maneuver with equal likelihood. The second baseline assumes that the agent vehicle is slowing down with 100\% probability, and thus always produces a plan to turn immediately. The first two baseline planners all use the RAO* planner to find the policy offline. Instead of using a risk-bounded planner, the third baseline planner chooses the action for the ego car based on the acceleration of the agent car. If the acceleration becomes smaller than a negative threshold, it assumes the agent car is slowing down and executes the ego car to make a left turn.

We test each planner with 1000 trials. Two measurements are used to quantify the performance in terms of safety and efficiency: the percentage of collision-free turns and the average time to make turns among all collision-free turns.

\begin{table}[t!]
\centering
\bgroup
\def\arraystretch{1.2}
\begin{tabular}{|l|l|l|}
\hline
    & \makecell{Success \\Rate [\%]} & \makecell{Completion \\Time [s]} \\ \hline
Conservative & \textbf{100.00} & 9.60   \\ \hline
Risky & 56.80 & \textbf{4.80}   \\ \hline
AccelerationBased & 82.10 & 8.87   \\ \hline
Ours ($\Delta=0.1\%$) & \textbf{100.00} & 8.64   \\ \hline \hline
Conservative & \textbf{100.00} & 26.74   \\ \hline
Risky & 28.60 & \textbf{19.20}   \\ \hline
Ours ($\Delta=0.1\%$) & \textbf{100.00} & 24.67   \\ \hline
Ours ($\Delta=1\%$) & 99.80 & 23.19   \\ \hline
\end{tabular}
\egroup
\caption{Performance of different intention-aware planners. Top half: unprotected left turn. Bottom half: lane change.}
\label{tbl:performance}
\end{table}

Top four rows in Table \ref{tbl:performance} summarize the performance of our planner and three baseline planners in the unprotected left turn scenario. The first baseline method is quite successful in avoiding all potential near collisions by using a conservative assumption. However, they sacrifice the efficiency by keeping the ego car waiting even when the agent car slows down and yields. On the other hand, the second baseline is efficient in making turns with the smallest amount of completion time, but leads to crashes almost half of the time because of its over-risky assumption. The acceleration-based method works in most of the cases, but it is vulnerable in examples where the agent car decelerates slightly in the middle of a forward maneuver. Compared to the methods above, our model is able to find safe plans in all cases by allowing only a small risk bound of 0.1\%, and completes the tasks more efficiently by detecting the slow-down intention of the agent car quickly and reliably.

\begin{figure}[t!]
\centering
\includegraphics[width=.9\columnwidth]{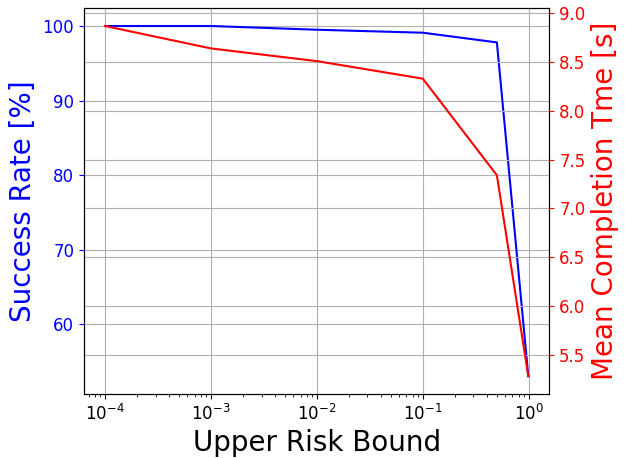}
\caption{The performance of our planner in the intersection left turn scenario as a function of the upper risk bound $\Delta$, where the $x$-axis is plotted in the log scale.}
\label{fig:left_turn_delta}
\end{figure}

We further analyze the performance of our planner by changing the risk bound $\Delta$. As plotted in Fig.~\ref{fig:left_turn_delta}, when the upper risk bound is relatively small, the planner can guarantee to find collision-free policies for the ego car because it is very rare that our estimator classifies a forward maneuver into a slow down maneuver for the oncoming agent car with great confidence. However, the smaller the risk bound, the more time it takes the ego car to complete the task as the planner needs more time to be certain that it is within the safe bound to make a left turn. When the risk bound becomes larger, the success rate starts to drop because in some cases the intent recognizer provides inaccurate estimations, which leads our planner that takes large risk bound to choose the wrong action for the ego vehicle. When the risk bound is set to one, the behavior of the planner is essentially identical to the second baseline planner that takes risky decisions.

Therefore, we can tune the \textit{style} of our planner using different risk bound values, where a small risk bound leads to safer behaviors, and a large risk bound produces a risky planner that is willing to sacrifice safety in order to achieve better efficiency. The planner style can be either determined offline by a user, or adjusted online during a planning episode based on different driving conditions. 

\subsection{Lane Change with Multiple Agent Vehicles} 
In this scenario, we aim to test our planner in a more complicated scenario, where the ego vehicle interacts with more than one agent car on a busy road and needs to make lane changes to achieve better efficiency. In addition to comparing the performances with baseline methods, we analyze the computational time of our planner by varying the number of agent cars and the planning horizon.

\subsubsection{Scenario Setup}
\begin{figure}[t!]
\centering
\includegraphics[width=.9\columnwidth]{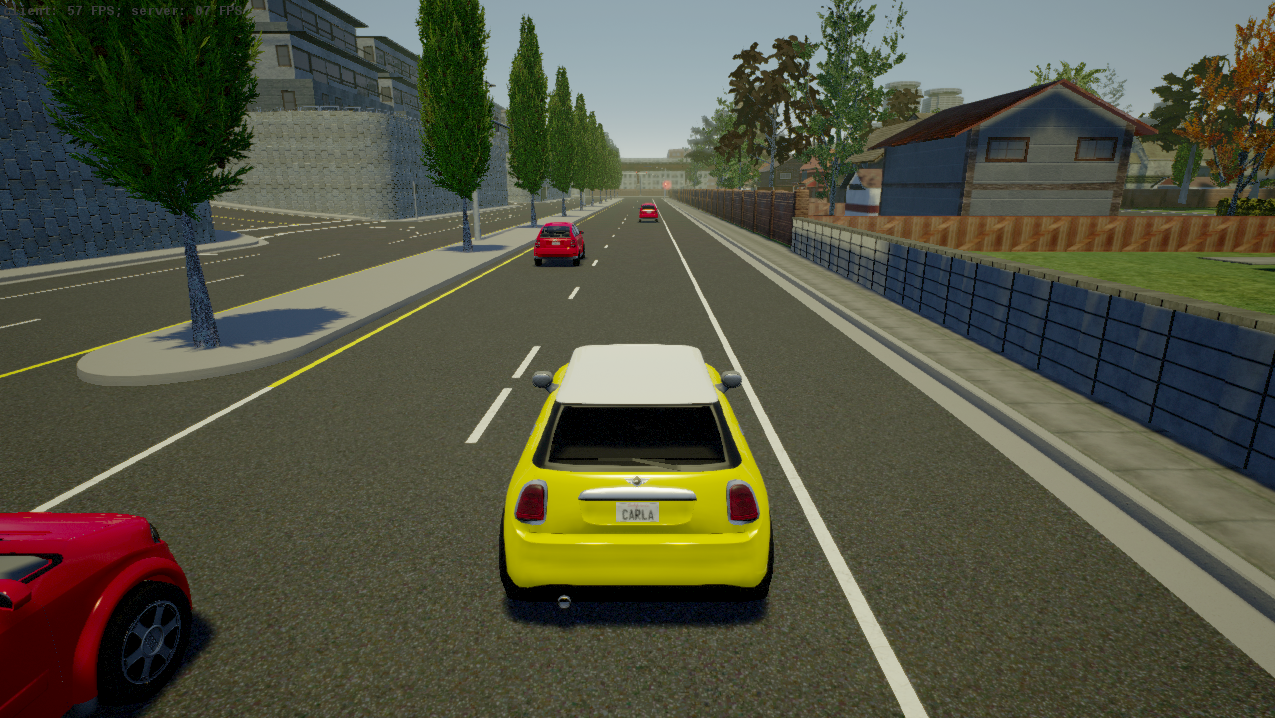}
\caption{A lane change test example, where the yellow vehicle is the ego car, and the reds are the agent cars.}
\label{fig:ego_lane_change}
\end{figure}

As shown in Fig.~\ref{fig:ego_lane_change}, we place the ego vehicle in the middle of a two-lane road at the beginning of each test. It can choose to change lane or go forward, and its goal is to reach the end of the road as soon as possible. In addition, we place three agent vehicles randomly next to the ego vehicle, each of which can choose to change lane or go forward. The agent cars are designed to move more slowly than the ego car, which encourages the ego car to bypass them by changing lanes. Again, a set of PFTs for all agent vehicle maneuvers are learned from human demonstrations, and we use the same $\epsilon$ value as in the previous scenario.

The starting location of each agent vehicle is chosen so that they can create potential risks to the ego car. Fig.~\ref{fig:ego_lane_change} shows one testing example, where an agent vehicle is placed behind the yellow ego car in the adjacent lane to prevent the ego car from making a lane change at the beginning of the test. The other two agent vehicles are placed in the front on different lanes to create potential risks. From many experiments, we find that three cars are sufficient to cover most dangerous lane change scenarios. 

\subsubsection{Results}

In this scenario, the intent recognition provides accurate results with an average maneuver estimation accuracy of 98.5\%, and an average trajectory prediction error of 2.23 meters. A better maneuver estimation performance is achieved compared to the left turn scenario as lane change and forward maneuvers can be more easily distinguished in the lateral axis. However, the trajectory prediction error increases because a misclassified change-lane maneuver will result in a trajectory prediction in the opposite direction.

Similar to the intersection left turn scenario, we test our system and two baseline planners that have simple prediction models with 1000 trials. The results are summarized in the bottom half of Table \ref{tbl:performance}. Our planner is able to produce more efficient plans than the conservative planner while maintaining safety all the time if the risk bound is set to be small. In addition, we can make an aggressive planner to further reduce the completion time by increasing the risk bound. Although the success rate drops, our planner is able to respect the risk bound with guarantees.

\begin{figure}[t!]
\centering
\includegraphics[width=.95\columnwidth]{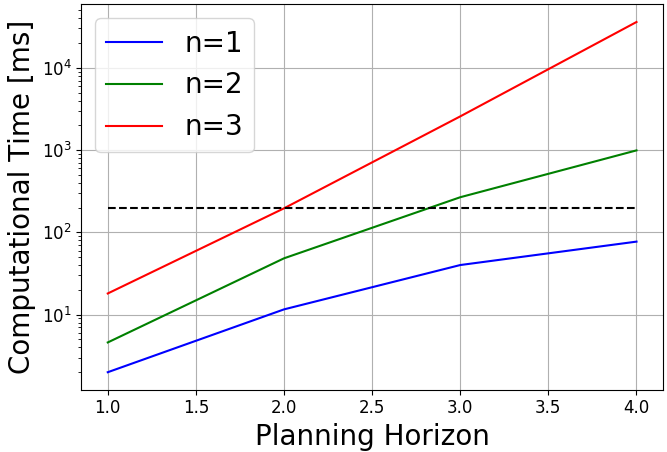}
\caption{Planning time using RAO* planner as a function of the planning horizon $H$ and the number of agent vehicles $n$. The planning time is shown in log scale.}
\label{fig:planning_time}
\end{figure}

We choose a planning horizon of two because the number of possible future states for the agent vehicles increases exponentially with the planning horizon, and the planner needs to compute the near collision risk for each possible future state. In Fig.~\ref{fig:planning_time}, we plot the planning time in log scale as a function of the planning horizon for different numbers of agent vehicles, where each vehicle can choose from three maneuvers and $\epsilon$ is fixed. The dashed black line indicates the maximum allowed planning time of 0.2 seconds if we want to keep the control frequency at 5Hz. With three agent vehicles and a planning horizon of two, it takes around 0.191 seconds for the planner to find a solution, which satisfies the planning time constraint. Although a planning horizon of two is not ideal, our ego vehicle is able to make adjustments quickly if there exists possible risk further ahead thanks to the online capability of our planner. Furthermore, we observe that in many cases the ego vehicle only needs to interact with one or two vehicles. Therefore, we can adjust the planning horizon based on the number of interacting agents to improve the performance of our planner.

\section{Conclusion}
\label{sec:conclusion}
In conclusion, we propose an online risk-bounded planning system that considers the intentions of surrounding dynamic agents, including their intended maneuvers and future motions. The intentions are recognized by a Bayesian filter based on a probabilistic hybrid automaton that models the agent vehicle behaviors. We integrate the intent recognition results into a POMDP model and solve it with a heuristic forward search method with upper bound guarantees on the near collision probability with other agent vehicles. Our system is demonstrated to work in two challenging dynamic environments in real time and outperform other baseline methods. We also show a further analysis of how the upper bound on risk constraint can change the behavior of our planner. We believe that both risk guarantee and intent recognition capability are necessary for safe-critical motion planning tasks in a dynamic environment, and our work provides a promising framework in that direction. One possible future work is to test our work in real systems and more complicated scenarios. In addition, we would like to apply our method to other domains such as interactive robotics manipulation. 

\fontsize{9.0pt}{10.0pt}
\bibliography{ref}
\bibliographystyle{aaai}
\end{document}